\documentclass[12pt]{article}

\usepackage[margin=0.7in]{geometry}     
\usepackage[parfill]{parskip}
\usepackage[utf8]{inputenc}
\usepackage[hidelinks]{hyperref} 

\usepackage{amsmath,amssymb,amsfonts,amsthm}
\usepackage{graphicx}
\usepackage{wrapfig}
\usepackage{paralist}

\title{The VIA Annotation Software for Images, Audio and Video}

\author{Abhishek Dutta \and Andrew Zisserman}
\date{Visual Geometry Group (VGG) \\ Department of Engineering Science\\ University of Oxford \\ \texttt{\{adutta,az\}@robots.ox.ac.uk}}

\begin{document}
\maketitle

\begin{abstract}
In this paper, we introduce a simple and standalone
manual annotation tool for images, audio and video: the VGG Image Annotator (VIA).
This is a light weight, standalone and offline software package that does not require
any installation or setup and runs solely in a web browser. The VIA software allows
human annotators to define and describe spatial regions
in images or video frames,  and temporal segments in audio or video.
These manual annotations can be exported to plain text data formats such as JSON
and CSV and therefore are amenable to further processing by other software tools.
VIA also supports collaborative annotation of a large dataset by a
group of human annotators. The BSD open source license of this software allows it
to be used in any academic project or commercial application.
\end{abstract}

\begin{figure*}
  \includegraphics[width=\textwidth]{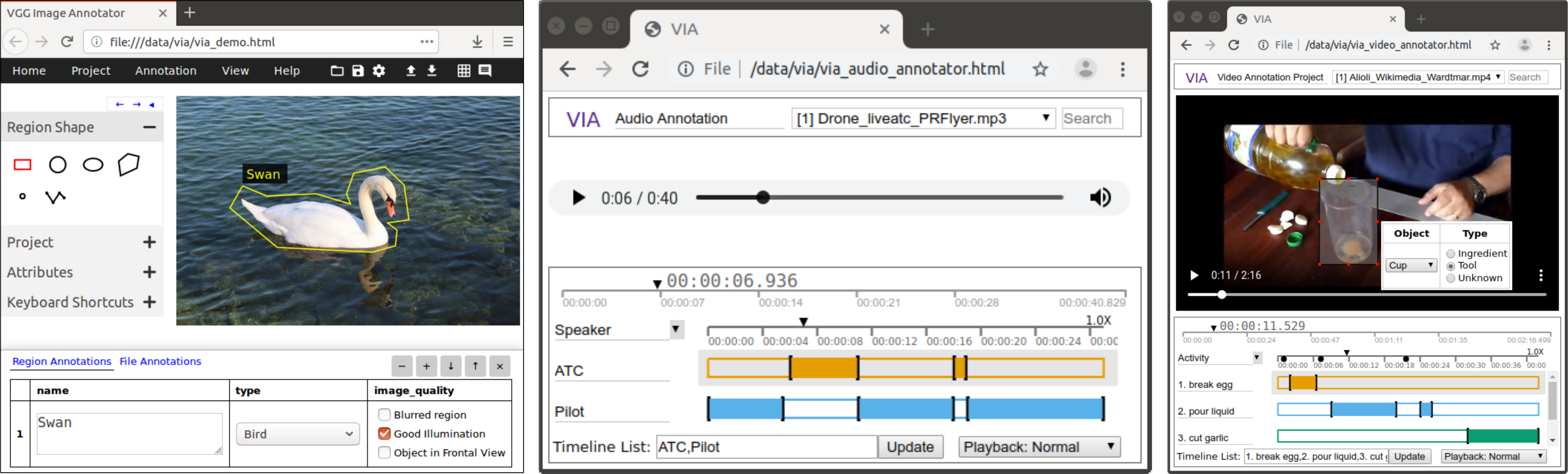}
  \caption{Screenshots of VIA running as an offline and standalone application
    in a web browser. (left) The image region occupied by a swan is defined using
    a polygon and described using the annotation editor. (middle) Speech
    segments of two individuals are manually delineated in an audio recording
    of a conversation between an Air Traffic Controller and pilot. (right) Temporal segments showing
    different human activities (e.g.\ break egg, pour liquid, etc.)
    and spatial regions (e.g.\ bounding box of cup) occupied by different objects
    in a still video frame are manually delineated in a video showing preparation
    of a drink.}
  \label{fig:via_teaser}
\end{figure*}

\section{Introduction}
Manual annotation of a digital image, audio or video is a fundamental
processing stage of many research projects and industrial applications.
It requires human annotators to define and describe spatial regions
associated with an image or still frame from a video, or delineate  temporal
segments associated with audio or video. Spatial regions are defined
using standard region shapes such as a rectangle, circle, ellipse,
point, polygon, polyline, freehand drawn mask, etc.\ while the temporal
segments are defined by start and end timestamps
(e.g.\ video segment from 3.1 sec. to 9.2 sec.).
These  spatial regions and temporal segments are described using textual
metadata.

In this paper, we present a simple and standalone manual annotation tool, the VGG Image
Annotator (VIA), that supports both spatial annotation of images and temporal annotation
of audio and videos. It is written using solely HTML, Javascript and CSS, and runs as an offline
application in most modern web browsers, without requiring any installation or setup.
The complete VIA software fits in a single self-contained HTML page of size less than $400$ kilobyte.
This light footprint of the VIA software allows it to be easily shared (e.g.\ using
email) and distributed amongst manual annotators. VIA can be downloaded from
\url{http://www.robots.ox.ac.uk/~vgg/software/via}.

Since VIA requires no installation or setup, and is up and running in a few seconds,
non-technical users can begin annotating their images, audio and video very quickly;
and consequently, we have seen widespread adoption of this software in a large
number of academic disciplines and industrial sectors. A minimalistic approach
to user interface design and rigorous testing
(both internally and by our vibrant open source community) has allowed the
VIA software to become an easily configurable, simple and easy to use manual
annotation tool.

The development of VIA software began in August 2016 and the first public
release of Version~1 was made in April 2017. Many new advanced features
for image annotation were introduced in Version~2 which was released in
June 2018. The recently released Version~3 supports
temporal annotation of audio and video. As of July 2019, the
VIA software has been used more than $1,000,000$ times ($+220,000$ unique pageviews).

This paper is organised as follows. We describe different use cases of the VIA
software in Sections~\ref{sec:image_annotation}--\ref{sec:audio_video_annotation}, and
the software design principles in Section~\ref{sec:software_design}.  A brief
overview of the open source ecosystem built around the VIA software
is included in Section~\ref{sec:open_source}.
The impact of VIA software on several academic
and industrial projects is discussed in Section~\ref{sec:impact}.
Finally, we describe our planned directions for extensions in Section~\ref{sec:conclusions}.

\section{Image Annotation}
\label{sec:image_annotation}
\begin{figure*}[h]
  \centering
  \includegraphics[width=\linewidth]{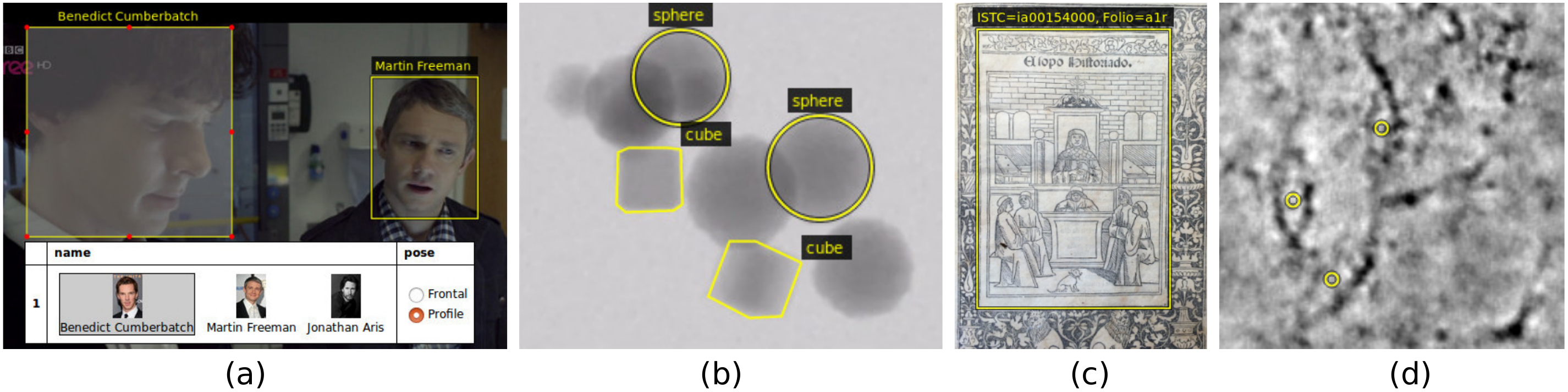}
  \caption{The VIA software is being used in a wide range of academic disciplines
    and industrial sectors to define and describe regions in an image.
    For example, (a) actor faces are annotated using rectangle shape and
    identified using a predefined list; (b) the boundary of arbitrarily shaped objects
    in scanning electron microscope image is defined using circle and polygon
    shapes by~\cite{bigparticle2018howto}, (c) 15th-century printed
    illustrations are annotated using rectangle shape~\cite{matilde2018a}, and (d) the point shape
    has been used by~\cite{brasch2019visualization}~to manually define the location of
    particles in cryo-electron microscopy image.}
  \label{fig:image_annotation}
\end{figure*}

The VIA software allows human annotators to define and describe regions in an image.
The manually defined regions can have one of the following six shapes: rectangle,
circle, ellipse, polygon, point and polyline. Rectangular shaped regions are
very common and are mostly used to define the bounding box of an object. Polygon
shaped regions are used to capture the boundary of objects having a complex shape.
The point shape is used to define feature points like facial landmarks, or
keypoints in MRI images, location of particles in microscopy images, etc. The
circle, ellipse and polyline shaped regions are less common but are essential
for some projects. Examples are given in ~\figurename~\ref{fig:image_annotation}.

The textual description of each region is essential for many projects. Such
textual descriptions often describe visual content of the region. While a plain
text input element is sufficient to update the textual description, VIA supports
the following additional input types with predefined lists:
checkbox, radio, image and dropdown.
The predefined list ensures label naming consistency between the human annotators.

\section{Image Group Annotation}
\label{sec:img_grp_annotation}
Annotation of large image datasets is rarely accomplished solely
by human annotators. More usually, a two stage process is used
to reduce the burden on human annotators:
\begin{inparaenum}[a)]
\item \textit{Automatic Annotation}: Computer vision algorithms are applied
  to the image dataset to perform a preliminary (but possibly imperfect) annotation of the images.
\item \textit{Manual Filtering, Selection and Update}: Human annotators review the
  annotations produced by the Automatic Annotation process and perform manual
  filtering, selection and update to retain only the good quality annotations.
\end{inparaenum}
This two stage process off-loads the burden of image annotation from human
annotators and only requires them to perform filtering, selection and update
of automatic annotations. The VIA software supports this two stage model of
image annotation using its \textit{Image Grid View} feature which is designed
to help human annotators filter, select and update metadata associated with a
group of images. The image groups are based on the metadata and regions defined
by automatic computer vision algorithms.

\begin{figure}[h]
  \centering
  \includegraphics[width=\linewidth]{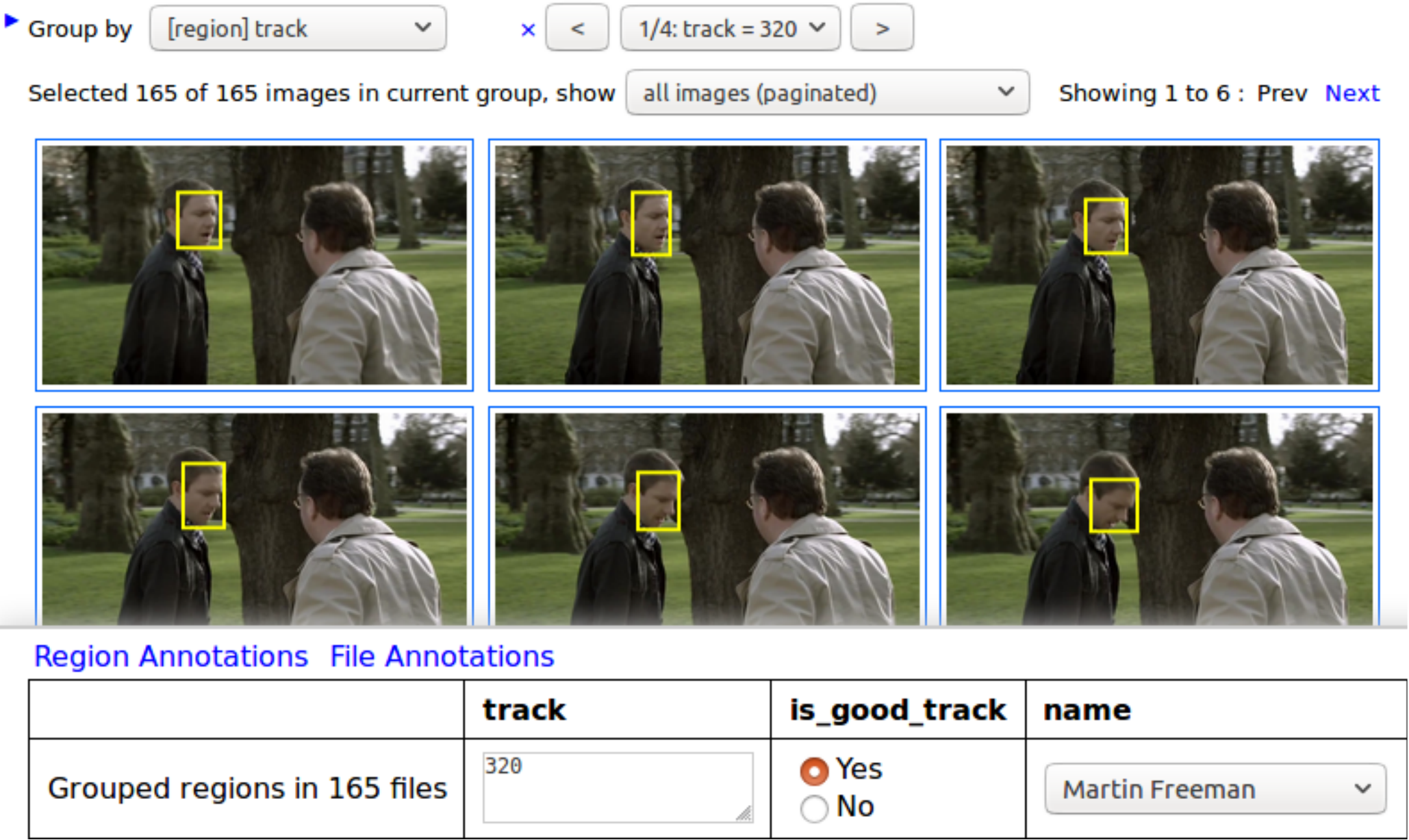}
  \caption{A set of automatically detected face tracks in $165$ consecutive video
  frames from BBC Sherlock series is assigned metadata (\texttt{is\_good\_track}
  and \texttt{name}) quickly by human annotators using the image grid view feature
  of VIA. A face track containing incorrect detections can also be easily filtered
  out by setting \texttt{is\_good\_track} to ``No''.}
  \label{fig:face_track_annotation}
\end{figure}

To illustrate the image grid view feature of VIA, consider the task of
face track annotation which involves delineating and identifying a face region of
an individual in consecutive frames of a video -- also called a face track. Such
annotated datasets are often used to train face detection and recognition algorithms.
For face track annotation, an automatic face detector (e.g.\ Faster R-CNN~\cite{ren2015faster})
is used to detect face regions in consecutive frames of a video and a face track
identification system (e.g.\ VGG Face Tracker~\cite{cao2019face}) identifies unique
face tracks from the automatically detected face regions in consecutive frames.
Automatically generated face track annotations are imported into the VIA software
and human annotators -- using the image grid view feature of VIA -- review these
automatically generated face tracks and select or filter the ones that are
correct as shown in~\figurename~\ref{fig:face_track_annotation}. Furthermore,
the VIA image grid view also allows bulk update of attributes associated with each
group. This capability of VIA allows human annotators to quickly annotate a large
number of images that have been partially annotated by automatic computer vision
algorithms.

The grid view also enables annotators to easily remove erroneous images from a group.
This functionality is very useful for re-training  an existing image classifier
by identifying images that have been incorrectly classified.

\section{Audio and Video Annotation}
\label{sec:audio_video_annotation}
\begin{figure}[h]
  \centering
  \includegraphics[width=\linewidth]{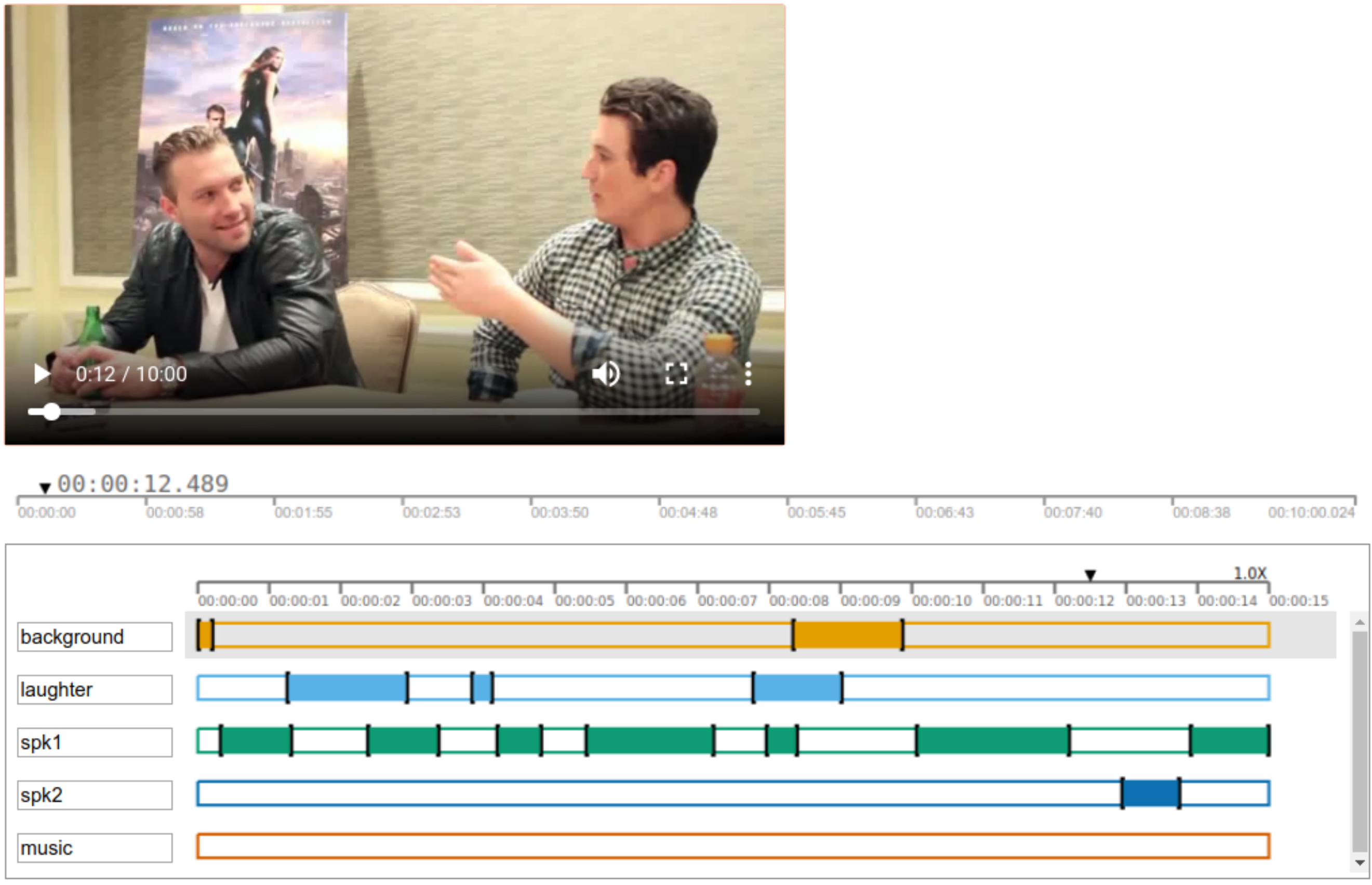}
  \caption{The VIA software being used to perform speaker diarisation for a video
    containing a conversation between two individuals. Human annotators manually identify the
    segments of the video that contains speech of an individual.}
  \label{fig:speaker_diarisation}
\end{figure}

The VIA software also allows human annotators to define temporal segments of an
audio or video and describe those segments using textual metadata. Such manually
annotated audio or video segments are useful for many projects. For instance, a large
number of human annotators are using VIA to collaboratively define temporal segments
containing speech of an individual (i.e.\ speaker diarisation) in videos taken
from~\cite{chung2018voxceleb2} as shown in~\figurename~\ref{fig:speaker_diarisation},
in order to assess the accuracy of automatic tools for
speaker diarisation. In a similar way, researchers
from Oxford Anthropology department are using VIA for identifying video segments
containing a particular chimpanzee in videos captured at a  forest site. The annotated
video segments are used to train and test computer vision algorithms that can
automatically detect and identify chimpanzees in videos captured in ``the wild''.
As an illustrative example for audio, we show the results of speaker diarisation on an audio
recording containing a conversation between an air traffic controller (ATC) and
a pilot in~\figurename~\ref{fig:via_teaser}~(middle).

\section{Software Design}
\label{sec:software_design}
The user interface of VIA is made using standard HTML components and
therefore the VIA software looks familiar to most new users. These
components are styled using CSS to achieve a greyscale colour scheme
which helps avoid distractions and focus attention on the visual content
that is being manually annotated using the VIA software. We follow the
minimalist approach for the user interface, and strive for simplicity
both in design and implementation. We resist adding new features or
updating existing user interface components if we feel that such change
leads to complexity in terms of usability and implementation. Most of our
design decisions are influenced by feedback from the open source community
thriving around the VIA software.

The HTML and CSS based user interface of VIA is powered by nearly $9000$ lines
of Javascript code which is based solely on standard features available in
modern web browsers. VIA does not depend on any external libraries. These design
decisions has helped us create a very light weight and feature rich manual
annotation software that can run on most modern web browsers without
requiring any installation or setup. The full VIA software sprouted from an
early prototype\footnote{\url{http://www.robots.ox.ac.uk/~vgg/software/via/via-0.0.1.txt}}
of VIA which implemented a minimal -- yet functional -- image annotation tool
using only $40$ lines of HTML/CSS/Javascript code that runs as
an offline application in most modern web browsers. This early prototype
provides a springboard for understanding the current codebase of VIA which
is just an extension of the early prototype. A detailed source code
documentation\footnote{\url{https://gitlab.com/vgg/via/blob/master/CodeDoc.md}}
is available for existing developers and potential contributors of the VIA open
source project.

Many existing manual annotation software (e.g.~\cite{russell2008labelme,zhang2018mask})
requires installation and setup. This requirement often results in a barrier for
non-technical users who cannot deal with variability in software installation
and setup procedure on different types of computing systems. The VIA software
and other recent annotation software tools like~\cite{pizenberg2018web} have overcome this
challenge by using the web browser as a platform for deployment of offline manual
annotation software. Since a standard web browser is already installed in most
computing systems, users can get up and running with our web browser based manual
annotation software in few seconds.

\section{Open Source Ecosystem}
\label{sec:open_source}
We have nurtured a large and thriving open source community which not
only provides feedback but also contributes code to add new features
and improve existing features in the VIA software. The open source
ecosystem of VIA is supported by its source code repository\footnote{\url{https://gitlab.com/vgg/via}}
hosted by the Gitlab platform. Most of our users report issues and
request new features for future releases using the issue portal. Many
of our users not only submit bug reports but also suggest a potential
fix for these software issues. Some of our users also contribute code
to add new features to the VIA software using the merge request portal. Thanks
to the flexibility provided by our BSD open source software license, many
representatives from commercial industry have contacted us through email to
seek advice for their engineering team tasked with adapting the VIA software for internal or commercial use.

\section{Impact on Academia and Industry}
\label{sec:impact}
The VIA software has quickly become an essential and invaluable research
support tool for many academic disciplines. For example, in Humanities, VIA has been used
to annotate hundreds of 15th-century printed illustrations~\cite{matilde2018a} and
annotate images ``which are meant to be read as texts''~\cite{pascoe2019scripttopict}.
In Computer Science, large numbers of image and video datasets have been manually annotated
by groups of human annotators using the VIA software~\cite{naphade2017nvidia}.
In the History of Art, VIA was used to manually annotate a multilayered
14th-century cosmological diagram containing many different elements~\cite{griffin2019diagram}.
In Physical Sciences, VIA is being used to annotate particles in electron
microscopy images~\cite{bigparticle2018howto,brasch2019visualization}. In Medicine, the VIA software
has allowed researchers to create manually annotated medical imaging
datasets~\cite{ferlaino2018towards,rakhlin2018neuromation,ali2019endoscopy}.

VIA has also been very popular in several industrial sectors which have
invested in adapting this open source software to their specific requirements. For
example, Puget Systems (USA), Larsen \& Toubro Infotech Ltd. (India) and Vidteq (Bangalore, India)
have integrated the VIA software in their internal work flow. Trimble
Inc.\  (Colorado, USA) adapted VIA for large scale collaborative annotation
by running VIA on the Amazon Mechanical Turk platform.

\section{Summary and Future Development}
\label{sec:conclusions}
In this paper, we described our manual annotation tool
called VGG Image Annotator (VIA). We continue to develop and
maintain this software according to the principles of open
source software development and maintenance.

VIA is a continually evolving open source project which aims to
be useful for manual annotation tasks in many academic
disciplines and industrial settings. This demands continuous
improvement and the introduction of advanced new features.
In future releases of the VIA software, we will introduce the following
features:
\begin{enumerate}
\item Collaborative Annotation: Annotating a large number of images
(e.g.\ a million images) or videos (e.g.\ thousands of hours of videos)
requires collaboration between a large number of human annotators.
Therefore, we are upgrading VIA to add support for collaborative annotation
which will allow multiple human annotators to incrementally and
independently annotate a large collection of images and videos.
A basic implementation of the collaborative annotation feature is already
available in the latest release of VIA.
\item Plugins: State-of-the-art computer vision models are becoming very
accurate for tasks such as segmenting and tracking  objects,
reading text, detecting keypoints on a human body,
and many other tasks commonly assigned to human annotators.
These computer vision models
can help speed up the manual annotation process by seeding
an image with automatically annotated regions and then letting human
annotators edit or update these regions to create the final annotation.
Thanks to projects like TensorFlow.js, it is
now possible to run many of these models in a web browser. We envisage such
computer vision models attached as plugins to VIA and running in the background
to assist human annotators. We are currently developing such a plugin to automatically
track an object in a video.
\end{enumerate}

\noindent \textbf{Acknowledgements}: This work is funded by EPSRC programme grant
Seebibyte: Visual Search for the Era of Big Data (EP/M013774/1).

\bibliographystyle{unsrt}
\bibliography{ref}

\end{document}